\pdfoutput=1

\documentclass[11pt]{article}

\usepackage{EMNLP2023}

\usepackage{times}
\usepackage{latexsym}

\usepackage[T1]{fontenc}

\usepackage[utf8]{inputenc}

\usepackage{microtype}

\usepackage{inconsolata}

\usepackage{subcaption}
\usepackage[nameinlink,capitalise]{cleveref}
\usepackage{graphicx}
\usepackage{booktabs}
\usepackage{longtable}
\usepackage{multicol}
\usepackage{multirow}
\usepackage{bm}
\usepackage{enumitem}
\usepackage{float}
\usepackage{xspace}
\usepackage{xcolor}         
\usepackage{colortbl}         
\usepackage{amssymb}  
\usepackage{url}
\newcommand{\subsetsumsplit}{\textsc{Subset-Sum-Split}}

\newcommand{\closestsplit}{\textsc{Closest-Split}}
\newcommand{\closestsplits}{\textsc{Closest-Splits}}

\title{Latent Feature-based Data Splits to Improve
Generalisation Evaluation: \\A Hate Speech Detection Case Study}

\author{Maike Z\"ufle$^1$ \and Verna Dankers$^1$ \and Ivan Titov$^{1,2}$\\
  $^1$ILCC, University of Edinburgh \\
  $^2$ILLC, University of Amsterdam \\
  \texttt{m.s.zufle@sms.ed.ac.uk \hspace{0.2cm}vdankers@ed.ac.uk \hspace{0.2cm}ititov@inf.ed.ac.uk}}

\begin{document}
\maketitle
\begin{abstract}
With the ever-growing presence of social media platforms comes the increased spread of harmful content and the need for robust hate speech detection systems.
Such systems easily overfit to specific targets and keywords, and evaluating them without considering distribution shifts that might occur between train and test data overestimates their benefit.
We challenge hate speech models via new train-test splits of existing datasets that rely on the clustering of models' hidden representations.
We present two split variants (\subsetsumsplit\ and \closestsplit) that, when applied to two datasets using four pretrained models, reveal how models catastrophically fail on blind spots in the latent space.
This result generalises when developing a split with one model and evaluating it on another.
Our analysis suggests that there is no clear surface-level property of the data split that correlates with the decreased performance, which underscores that task difficulty is not always humanly interpretable.
We recommend incorporating latent feature-based splits in model development and release two splits via the GenBench benchmark.\footnote{Our implementation is available at \url{https://github.com/MaikeZuefle/Latent-Feature-Splits}}

\end{abstract}

\section{Introduction}\label{sec:introduction}

Developing generalisable hate speech detection systems is of utmost importance due to the environment in which they are deployed.
Social media usage is rapidly increasing, and the detection of harmful content is challenged by non-standard language use, implicitly expressed hatred, a lack of consensus on what constitutes hateful content, and the lack of high-quality training data \citep{yin2021towards}.
When developing hate speech detection models in the lab, it is, therefore, vital to simulate evaluation scenarios requiring models to generalise outside the training context.
`In the wild', NLP models may encounter text from different periods \citep{lazaridou-et-al-2021-temporal}, authors \citep{huang-paul-2019-neural} or dialects \citep{ziems-etal-2022-value}, including unseen words \citep{elangovan-etal-2021-memorization} and words whose spelling changed or was obfuscated \citep{serra2017class}.
Performing successfully on this data despite such distributional changes is called \textit{out-of-distribution} (o.o.d.) generalisation. 

\begin{figure}
    \centering
    \includegraphics[width=0.4 \textwidth]{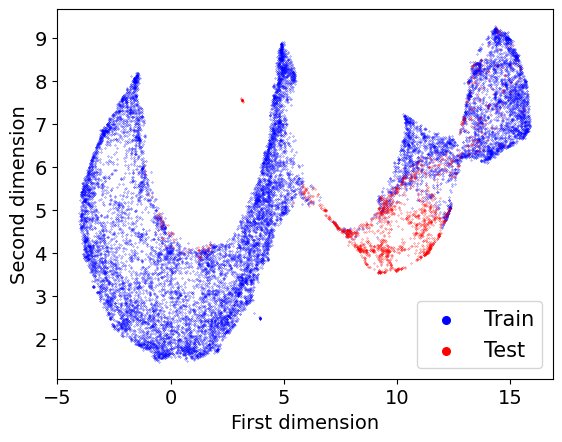}
    \caption{A UMAP projection of BERT's representations, showing the proposed train-test split, that is constructed by grouping clusters in the latent space.}
    \label{fig:data_split_into}
    \vspace{-0.5cm}
\end{figure}

How can the ability to generalise best be measured? Despite recent work illustrating that i.i.d.\ testing does not adequately reflect models' generalisability \citep[e.g.][]{sogaard-etal-2021-need}, evaluation using randomly sampled test sets is still the status quo \citep{rajpurkar-etal-2016-squad,wang-2018-glue,wang-2019-superglue,muennighoff-etal-2023-mteb}. Potentially, this is because obtaining and annotating new data is expensive, and it is hard to define what o.o.d.\ data is \citep{arora-etal-2021-types}. For \textit{humans}, properties like input length \citep{varis-bojar-2021-sequence} or spelling mistakes \citep{ebrahimi-etal-2018-hotflip} might determine difficulty. But this need not be the same for \textit{models}. 
Evaluating models using a notion of model-dependent difficulty is gaining some traction \citep[e.g.][]{godbole2022benchmarking} but still remains largely unexplored. 

Contributing to that line of work, we propose a method that reuses existing datasets but splits them in a new way by relying on models' latent features.
We cluster hidden representations using $k$-means and distribute clusters over the train and test set to create a data split.
An illustrative example of such a split is shown in \cref{fig:data_split_into}. We present two variants (\subsetsumsplit\ and \closestsplit). While this method is in principle applicable to any classification problem, we experiment with four language models and two hate speech datasets (that include Reddit, Twitter and Gab data).
The results suggest that these splits approximate worst-case performance.
Models fail catastrophically on the new test sets, while their performance on independent test data is on par with other systems trained on i.i.d.\ training sets.
The difficulty is relatively stable across different models.
We analyse the data splits through correlation analyses, and do not find one clear surface-level property of the data split to be predictive of split difficulty. This underscores that model-based difficulty can be quite elusive.
We release two of our data splits for inclusion in the GenBench benchmark.

The remainder of this work is structured as follows: \cref{sec:related_work} elaborates on related work, followed by the introduction of the hate speech datasets (\cref{sec:data}) and the proposed splitting method (\cref{sec:methods}).
\cref{sec:results} presents model evaluation results, \cref{sec:analysis} analyses the splits in detail, and we conclude in \cref{sec:conclusion}. The GenBench eval card can be found in \cref{app:eval_card}.

\section{Related Work}\label{sec:related_work}

This section discusses related work on o.o.d.\ generalisation evaluation (\cref{subsec:gen_eval_rl}), followed by a discussion on why generalisation is a persisting challenge in hate speech detection (\cref{subsec:hate_speech_rl}).

\subsection{Generalisation evaluation}
\label{subsec:gen_eval_rl}
It is now well-established within NLP that models with high or even human-like scores \citep[e.g.][]{chowdhery2022palm} on i.i.d.\ splits do not generalise as robustly as the results would suggest. This has been demonstrated using synthetic data \citep[i.a.][]{lake-2017-systematic, mccoy-etal-2019-right, kim-linzen-2020-cogs} and for natural language tasks \citep[i.a.][]{sinha-etal-2021-masked, sogaard-etal-2021-need, razeghi-etal-2022-impact}. 
Alternative methods of evaluation have become more prominent, such as testing with different domains \citep[e.g.][]{tan-etal-2019-domain, kamath-etal-2020-selective, yang2022gluex} and adversarial testing, using both human-written \citep{kiela-etal-2021-dynabench} and automatically generated adversarial examples \citep[e.g.][]{zhang-2020-adversarial, chen-etal-2019-codah,gururangan-etal-2018-annotation,ebrahimi-etal-2018-hotflip}.

However, these types of evaluation require collecting or creating new data points, which is not always feasible for datasets that have been in use for years.
Re-splitting existing datasets in a non-i.i.d.\ manner makes more efficient use of existing datasets, 
and, accordingly, new data splits have been developed, that typically use a feature of the input or the output to separate train from test examples.
Splits that rely on the input use, for example, word overlap \citep{elangovan-etal-2021-memorization}, linguistic structures \citep{sogaard-2020-languages}, the timestamp \citep{lazaridou-et-al-2021-temporal}, or the context of words in the data \citep{keysers-et-al-2019-measuring} to generate a split.
Similarly, \citet{broscheit-etal-2022-distributionally} maximise the Wasserstein distances of train and test examples. 
Alternatively, one can evaluate generalisation using output-based non-i.i.d.\ splits: \citet{naik-etal-2018-stress} analyse the predictions of a model to find challenging phenomena, and  \citet{godbole2022benchmarking} re-split a dataset based on the predicted log-likelihood for each example.

The splitting method we propose relies neither on the discrete input tokens nor the output, but instead uses the internal representations of finetuned models.

\subsection{Hate speech detection}
\label{subsec:hate_speech_rl}

With the rise of social media platforms, hate speech detection gained traction as a computational task \citep{jahan-2023-survey}, leading to a wide range of benchmark datasets. Most of these datasets rely on data from social media platforms, such as Reddit \citep{qian-etal-2019-benchmark, vidgen-etal-2021-introducing}, Twitter \citep{elsherief-etal-2021-latent}, Gab \citep{qian-etal-2019-benchmark,mathew-2020-hatexplain}, or Stormfront \citep{gibert-2018-hate}. 
This work is restricted to hate speech classification using a Reddit dataset \citep{qian-etal-2019-benchmark} and a Twitter and Gab dataset \citep{mathew-2020-hatexplain}, which we will elaborate on in \cref{sec:data}. 

Recent advances in NLP such as the introduction of large language models have led to impressive results in hate speech detection \citep{fortuna-2019-survey, vidgen-etal-2019-challenges}. 
Nonetheless, non-i.i.d.\ generalisation is a persisting challenge \citep{yin-2021-generalisable}, because models tend to overfit to specific topics \citep{nejadgholi-kiritchenko-2020-cross, bourgeade-etal-2023-learn}, social media users \citep{arango-2019-general}, or keywords, such as slurs or pejorative terms \citep{dixon-2018-measuring, kennedy-etal-2020-contextualizing, waseem-2018-bridging, Palmer_Carr_Robinson_Sanders_2020, kurrek-etal-2020-towards}. When such overt terms are missing, models often fail to detect hate speech \citep{elsherief-etal-2021-latent}. 
In response to these generalisation issues, recent works combine existing hate speech datasets \citep{fortuna-2018-merge, Salminen-2020-combine, Chiril-2022-multitarget, bourgeade-etal-2023-learn}, which is a challenging task in itself considering the inconsistent definition of hate-speech across datasets \citep{nejadgholi-kiritchenko-2020-cross}.

Augmenting datasets or evaluating whether a model overfits to particular users or data sources requires annotated data.
However, these characteristics are often unavailable due to privacy requirements or because the annotations were not included in the dataset release.
Therefore, this work aims to find a data split that can evaluate generalisation without such annotations, relying instead only on a model's internal representations.

\section{Data}\label{sec:data}
We develop and evaluate our splitting method using the following two hate speech datasets.

\subsection{Reddit}
We use a widely used topic-generic Reddit dataset, proposed by \citet{qian-etal-2019-benchmark}. The dataset includes 22,317 examples. 
Each example in the dataset is labelled as either \textit{hate} (23.5\%) or \textit{noHate} (76.5\%). The dataset was collected from ten different subreddits by retrieving potential hate speech posts using hate keywords taken from \citet{ElSherief_Nilizadeh_Nguyen_Vigna_Belding_2018}.
The hate keywords correspond roughly to the following categories: 
\textit{archaic}, \textit{class}, \textit{disability}, \textit{ethnicity}, \textit{gender},
\textit{nationality}, \textit{religion}, and \textit{sexual orientation}.
The data is structured in conversations that consist of at most 20 comments by the same or different authors. These comments were manually annotated with \textit{hate} or \textit{noHate}, with each annotator assigned five conversations.  

\subsection{HateXplain}
The second dataset is HateXplain \citep{mathew-2020-hatexplain}, which is also topic-generic and widely used. It contains 20,148 examples from Twitter and Gab. Posts from the combined collection were filtered based on a lexicon of hate keywords and phrases by \citet{Davidson_Warmsley_Macy_Weber_2017, mathew-2019-spread, ousidhoum-etal-2019-multilingual}. The selected posts were then manually annotated.
HateXplain examples are labelled as either \textit{hateful} (31\%), \textit{offensive} (29\%) or \textit{normal} (40\%), as proposed by \citet{Davidson_Warmsley_Macy_Weber_2017}.
Offensive speech differs from hate speech in that it uses offensive terms without directing them against any person or group in particular. 
All offensive and hate examples are annotated with the community that they target. These communities include, among others, \textit{Africans}, \textit{Jewish People}, \textit{Homosexuals} and \textit{Women}, and we use them for further analysis of our data splits in \cref{sec:analysis}.

\section{Methodology}\label{sec:methods}
\begin{figure}
    \centering
    \includegraphics[width=\columnwidth]{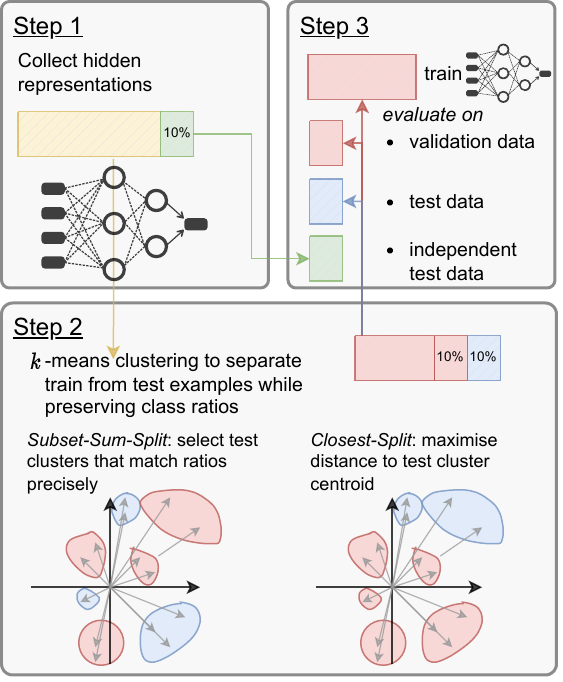}
    \caption{Overview of the proposed splitting method.}
    \label{fig:method_overview}
\end{figure}

Our proposed splitting strategy, for which we introduce two variants, is detailed in \cref{subsec:method_construct}.
We evaluate our splits through comparisons to a random splitting baseline and on external test sets. We discuss the corresponding experimental setups in
\cref{sec:experiments}.

\subsection{Constructing Data Splits}\label{subsec:method_construct}

The construction of the data splits involves three steps, that are depicted in \cref{fig:method_overview}. In step 1, the method extracts the latent representations of inputs from a language model that was finetuned on the task using one of the hate speech datasets mentioned above. In step 2, the data is clustered based on these representations and clusters are assigned to either the train or the test set. In step 3, language models are then trained and evaluated on this new split. In addition to the obtained test set, the language models are also evaluated on independent test data, that was set aside for this purpose.\footnote{Note that the split thus only includes 90\% of the data. Setting aside the 10\% is for quality control of the models and could be omitted when future work applies our method.}

The key idea behind the approach is  that language models implicitly capture salient features of the input in their hidden representations, where inputs with similar properties are close together \citep{thompson-2020-topic, grootendorst2022bertopic}. Assigning clusters to the train and test set thus accomplishes separation based on latent features, and by finetuning we ensure that the clusters separate examples based on \textit{task-specific} features. 

\paragraph{Obtaining Hidden Representations}
We finetune a language model for the given task, using the independent test data as validation set to optimise hyperparameters. We then obtain latent representations for each input example, leveraging the 
representation of the \texttt{[CLS]} token after the final layer as a representation of the input, as is commonly done \citep[e.g.][]{may-etal-2019-measuring, qiao-2019-understanding}. 

Since for high-dimensional data, distance metrics fail to accurately capture the concept of proximity \citep{beyer-1999-dims, aggarwal-2001-dims} and tend to overly rely on individual dimensions
\citep{timkey-van-schijndel-2021-bark} we conduct experiments with low-dimensional representations and full-dimensional ones. To this end, we either project the full representations 
into $d_{U}$-dimensional spaces using UMAP post-training \citep{mcinnes2020umap}, 
or obtain $d_{B}$-dimensional representations by introducing a bottleneck in the model between the last hidden layer and the classification layer. The bottleneck is a linear layer that compresses the hidden representations, forcing the model to encode the most salient latent features into a low-dimensional space before classifying the examples.

\paragraph{Clustering and Splitting the Data}
Each representation from step 1 gives the position of an input example in the latent space. The examples are clustered in this space using the $k$-means algorithm \citep{lloyd-1982-kmeans}.

Hyperparameters of the $k$-means clustering can be found in \cref{tab:k-means}.
After clustering, each cluster is assigned to either the train or the test set, keeping two constraints: A fixed test data size (we choose 10\%) and train and test set need to have equal class distributions.
Without equal class distributions, it would be unclear whether changes in performance are due to the increased difficulty of the test set, or the changes in label imbalance.
A partition of the dataset that fulfils these constraints will be referred to as \textit{target} in this work.

To reach the target test set, two algorithms, \subsetsumsplit\ and \closestsplit, are designed to decide how to split the clusters. 
Both algorithms lead to an under-representation of parts of the latent space in the model's training set, but whilst \subsetsumsplit\ might under-represent smaller, potentially distant pockets of the latent space, \closestsplit\ under-represents a single connected region.
The algorithms are explained in detail below.

\paragraph{Method 1: \subsetsumsplit}
The constraints on the class and test ratios explained above, and the additional constraint of keeping whole clusters together can be described by the Subset Sum Problem \citep{Kellerer2004}. In this setting, the Subset Sum Problem can be modified to a \textit{multidimensional} Subset Sum Problem: The multidimensional target consists of the number of desired test examples for each class in the dataset. The task is then to select a subset of the clusters, such that the number of examples for each class sums up to the desired target. 
To improve the chances of reaching the desired target, the Subset Sum Problem is solved for $k=3$ to $k=50$ clusters and the solution closest to the desired target using the smallest $k$ is taken as the test set.  If the closest solution does not match the exact target sum, examples from another randomly selected cluster are used to complete the test set.
Note that the clusters in the test set do not necessarily lie close to each other in the latent space, as this is not a constraint for this algorithm. 

\paragraph{Method 2: \closestsplit}
In contrast to the \subsetsumsplit, the \closestsplit\ aims to put as much distance as possible between the train and test clusters. This leads to an even bigger under-representation of parts of the latent space in the training set. 
Once the clusters have been computed, their centroids are calculated. The cluster that lies farthest away from all the other clusters is identified and added to the test set. If the size of the farthest cluster exceeds the target test set size, the next farthest cluster is taken instead.
Cosine similarity between cluster centroids is used as the distance measure.  Then  \textit{nearest neighbour} clustering with the cluster centroids is performed, as long as the size of the test set does not exceed the target size. When this nearest-neighbour clustering is finished, individual examples that are closest to one of the test set centroids are added to the test set until the target size is reached.
As for the \subsetsumsplit, the algorithm is performed for $k=3$ to $k=50$ clusters.
$k$ is selected such that the number of individual examples added is minimised.

\subsection{Evaluating Splits' Difficulty}\label{sec:experiments}

\paragraph{Models} We use four transformer language models to obtain and evaluate the data splits: BERT-Base(-Cased)  \citep{devlin-etal-2019-bert}, its smaller variant  BERT-Medium \citep{turc-2019-medbert, 
bhargava2021generalization}, HateBERT \citep{caselli-etal-2021-hatebert}, a BERT-Base-Uncased model that was further pretrained on abusive Reddit data using the MLM objective, and RoBERTa-Base \citep{liu-2019-roberta}.
From these models, we extract the full hidden representations, hidden representations via a bottleneck, for $d_{B}\in\{10, 50, 200\}$, and hidden representations post-processed using UMAP, for $d_{U}\in\{10, 50, 200\}$.

\paragraph{Model Evaluation}
Having obtained data splits based on four language models and hidden dimensions with different sizes, the first way of evaluating models is by finetuning the language models on their respective \subsetsumsplit\ and \closestsplit. The hyperparameters used for finetuning are listed in \cref{tab:lm_hyperparameters}, \cref{app:clusterh}, and we estimate $d_U$ and $d_B$ by varying their values for the Reddit dataset. We compare the results obtained with the proposed data splits to a baseline split, which takes the same examples but splits them randomly while maintaining class proportions.
Random splits are generated using three different seeds, and the proposed data splits are obtained with three different clustering seeds. For each data split involved, the models are trained with three seeds that determine the classifier's initialisation and the presentation order of the data. The results are averaged accordingly.

The evaluation metrics are accuracy and F1-scores. For the Reddit dataset, the F1-score is the score of the \textit{hate} class, whereas for HateXplain, the F1-score is macro-averaged over the three classes.

To better understand the robustness of the results, we perform an additional set of experiments on the most challenging data splits observed, to answer the following questions:
\begin{enumerate}[noitemsep, leftmargin=*,topsep=0pt]
    \item \textit{Is split difficulty driven by the input or by task-specific latent features?} For the Reddit data, we split the dataset based on task-agnostic hidden representations obtained from pretrained models to analyse whether task-specific representations (i.e. representations finetuned on the task) are needed to create challenging data splits. 
    \item \textit{Do models trained on new splits perform on par with conventional models on independent data?} Using HateXplain, we test the finetuned models on the independent test data that was set aside earlier to ensure that the newly obtained train data is still informative enough for test data sampled according to the original distribution.
    \item \textit{Is the difficulty of the data splits model-independent?} We also examine whether a split obtained by the hidden representations of a specific model is also challenging for other models using HateXplain data.
\end{enumerate}

\section{Results}\label{sec:results}
\begin{figure*}[ht]
\centering
\begin{subfigure}[b]{0.45\textwidth}
    \includegraphics[width=0.933 \linewidth]{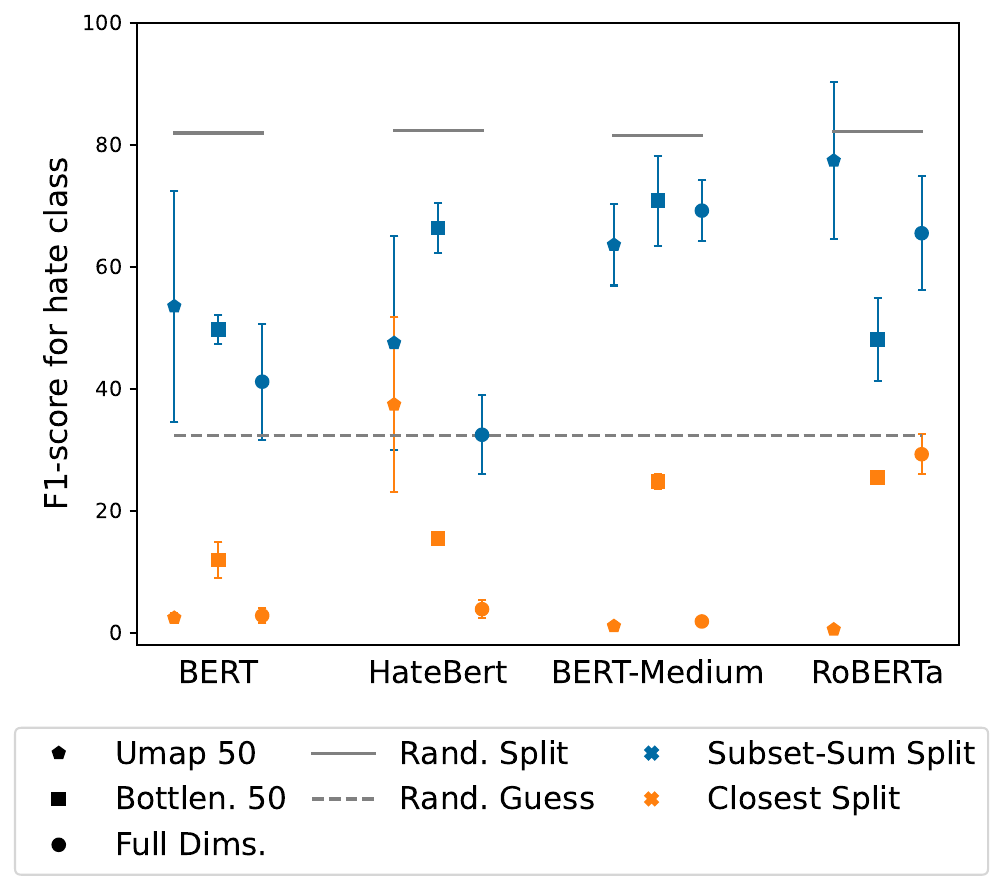}
    \caption{Reddit Dataset} 
    \label{fig:reddit_f1}
\end{subfigure}
\begin{subfigure}[b]{0.45\textwidth}
    \includegraphics[width=0.933\linewidth]{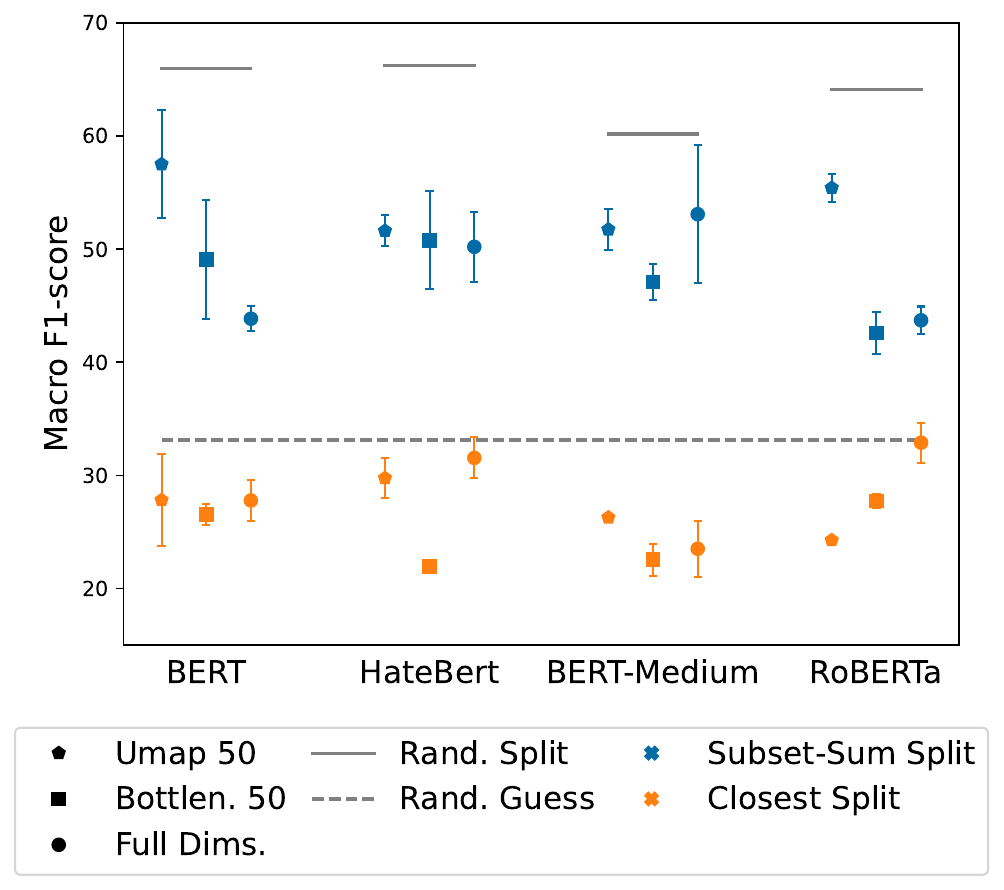}
   \caption{HateXplain Dataset}
    \label{fig:hatexplain_f1}
\end{subfigure}
\caption{Performance of models trained on the \subsetsumsplit\ and \closestsplit\ . The errorbars show the standard error between cluster seeds.
Horizontal lines indicate performance for models trained and tested on a random split. }
\label{fig:f1_drops}
\vspace{-0.3cm}
\end{figure*}

We now turn to evaluating models' performance on our newly proposed splits.

\subsection{Performance on Challenging Splits}\label{subsec:main_results}
\begin{table}
    \centering
\resizebox{\linewidth}{!}{\begin{tabular}{lcc}
\toprule
  model   &     Reddit Hate F1      &  HateXplain Macro F1          \\
\midrule
 BERT-base &  81.96 $\pm$ 0.5 & 66.0 $\pm$ 0.36  \\
BERT-medium   & 81.58 $\pm$ 0.66 &  60.18 $\pm$ 0.42  \\  
 HateBert & 82.34 $\pm$ 0.59 & 66.25 $\pm$ 0.35\\
  RoBERTa    &   82.15 $\pm$ 0.61 &  64.1 $\pm$ 0.9 \\   
\bottomrule
\end{tabular}}
    \caption{Results for the Reddit and HateXplain dataset on random splits using 90\% of the data. Random splits are generated using three different seeds and models are trained with three initialisation seeds. Mean and standard errors are reported.}
    \label{tab:baselines_random_split}
    \vspace{-0.3cm}
\end{table} 
We compare the performance of models trained on a random split to models trained on the \closestsplit\ and \subsetsumsplit. The random split performances are presented in \cref{tab:baselines_random_split}.  
For the binary Reddit dataset, performance on random splits is high for all four models with F1-scores for the hate class of around $82\%$.
The performance on the three-way HateXplain dataset is comparably lower, with macro F1-scores of around $65\%$.
For both datasets, these results are on par with (or surpass) baselines from prior work, upon which we elaborate in \cref{app:baselines}.\footnote{Note that these results are obtained with 90\% of the data as explained in \cref{sec:experiments}. The reader is referred to \cref{tab:reddit_baselines} and \cref{tab:hatexplain_baselines}  for accuracy results, results on 100\% of the data and results on the standard split.}

\paragraph{Hyperparameter Estimation} For both splits, we conduct a hyperparameter estimation to select $d_{U}$ and $d_{B}$ using the Reddit dataset, for which the results are shown in \cref{fig:reddit_results_more_dims}, \cref{app:split_hyper}. Across the board, all values considered challenge the models more than the random split, but full dimensions, $d_{U}=50$ and $d_{B}=50$ lead to a large decrease with relatively small variance between cluster seeds.

In addition to varying the dimensionalities, we consider using the models' pretrained representations (without further finetuning) to examine whether the latent features must be task-specific to challenge our models. 
Task-specific representations are, indeed, vital, as is shown in \cref{fig:reddit_pretrained_f1}, \cref{app:split_hyper}.

\paragraph{New Data Splits Reveal Catastrophic Failure}
Both \subsetsumsplit\ and \closestsplit\ lead to an under-representation of parts of the latent space in the model's training set and we hypothesised that this leads to a challenging data split. Indeed, the empirical results show significant performance drops when training models on these splits in comparison to random splits.

\cref{fig:reddit_f1} shows the performance drops for the Reddit dataset. For the \subsetsumsplit, F1-scores for the hate class drop significantly for all four models, but with a high variation between different cluster seeds. 
For the \closestsplit, test set performance drops even further and more consistently without much variation between cluster seeds: F1-scores for the hate class are mostly between $0$ and $25\%$.
\footnote{These results are not specific to the examination of F1-scores; the same tendencies can be observed when looking at the accuracy (\cref{app:accs_splits}).} 

\cref{fig:hatexplain_f1} displays performances for HateXplain, which similarly shows a drop in performance for 
\subsetsumsplit\ and  \closestsplit. \closestsplit\ leads to F1-scores that are on par with or below random guessing, resulting from drops of around $36\%$.

Overall, the \closestsplit\ is more challenging than the \subsetsumsplit.
Moreover, the bottleneck-based splits generally lead to the most stable results, i.e., the variance between different cluster seeds is the lowest.
In some cases performance drops below the random guessing baseline; this happens when a model fails to predict some class completely, defaulting instead to one of the other classes.
In summary, the new splits lead to drastic performance drops for both datasets and across all four models.

\subsection{Independent Test Set Performance}
We now take the most challenging split observed (\closestsplit\ with $d_B=50$) and further analyse the behaviour of models trained on this split for the HateXplain dataset, which is the most widely used dataset as well as the most challenging one.

\begin{figure}
    \centering
    \includegraphics[width=0.93 \linewidth]{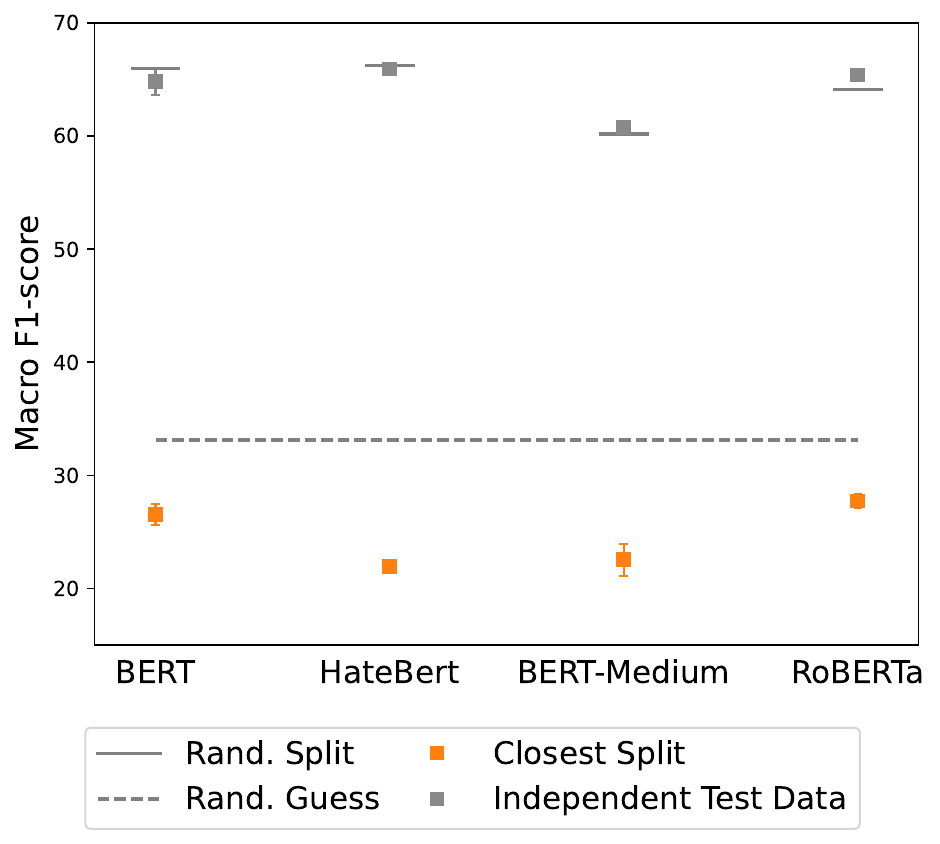}
    \caption{Performance of models trained on training data determined by the \closestsplit\ and evaluated on the test data of the \closestsplit\ and on independent test data (HateXplain dataset). Horizontal lines indicate performance for models trained and tested on a random split. 
    Errorbars show the standard error between cluster seeds.}
    \label{fig:hatexplain_ind_test_f1}
    \vspace{-0.3cm}
\end{figure}

From the results in \cref{subsec:main_results} it is clear that \closestsplit\ reveals weaknesses in these models, since the models struggle to generalise to the split's test data.
The question remains whether the test set obtained by the new splitting methods is harder or whether the new splitting method leads to very simple or perhaps even incomplete training sets, thereby preventing the models from learning the task. To this end, we evaluate the models trained on the training data obtained from a \closestsplit\ on the 10\% independent test data that was set aside earlier (\cref{subsec:method_construct}). The results show that models achieve similar performance on the independent test data as the models trained and tested on random data, strengthening the hypothesis that \closestsplit\ training data is informative enough to learn the task. Results for these experiments are reported in \cref{fig:hatexplain_ind_test_f1}.\footnote{The validation accuracy for the models trained on \closestsplit\ is for most splits around $5$ points higher than the accuracy on the validation set of the random data split---i.e.\ the models perform normally during training as suggested by the validation data.}

\subsection{Cross-Model Generalisation}\label{sec:cross_model}
\begin{figure}
    \centering
    \includegraphics[width=0.85\linewidth]{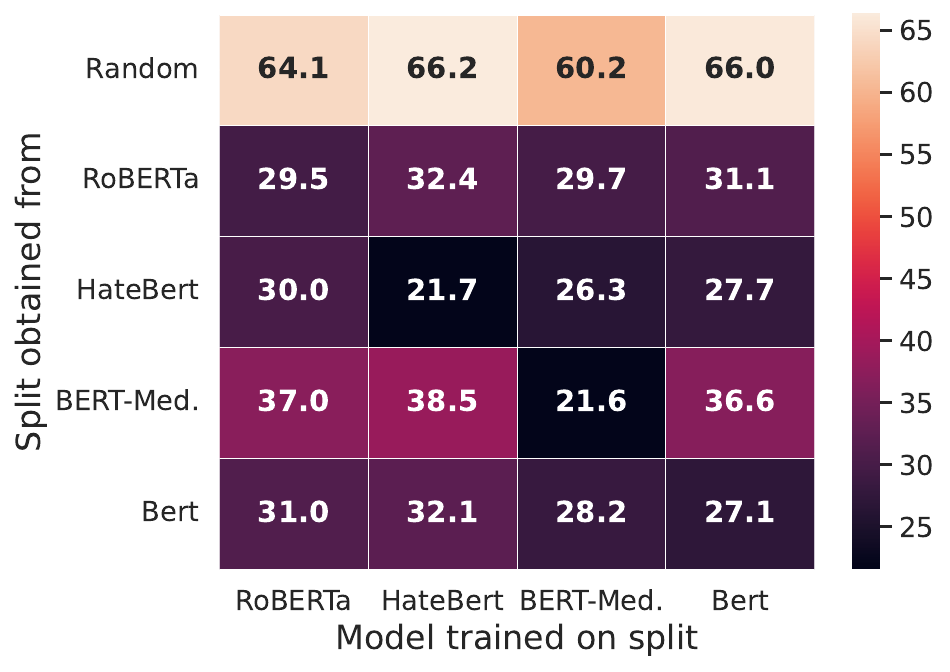}
    \caption{F1-scores for HateXplain on a \closestsplit\ ($d_B=50$). Comparison of models trained on the data split obtained with their respective hidden representations (diagonal) and on data splits obtained from representations of other models.}
    
    \label{fig:hatexplain_crossmodel}
    \vspace{-0.3cm}
\end{figure}

The previous results have shown that \closestsplit\ leads to challenging test sets. To show the robustness of these splits, we also examine whether these test sets are generally difficult or only for the model used to develop the split---i.e.\ we examine cross-model generalisation. 
The results of the cross-model evaluations can be seen in \cref{fig:hatexplain_crossmodel}. 
They show that data splits developed using one model are indeed also challenging for other models, although the personalised splits are slightly more challenging. 
These results do not only strengthen the robustness of the challenging data split, but have also practical implications: The data-splitting pipeline only needs to be carried out with one model and multiple models can be assessed and compared with the same split.

\section{Analysis}\label{sec:analysis}
The performance of models deteriorates heavily when using the proposed splits. This section analyses the generated splits; first examining the surface-level properties of the resulting train and test sets, and then taking a closer look at two specific splits by visualising the datapoints in the train and test sets. Additionally, an analysis of the topics in the train and test sets can be found in \cref{app:topics}.

\subsection{Correlation Analysis: Relating Splits' Features to Performance Drop}

For the most challenging split variant, \closestsplit, we investigate the correlation of performance drops compared to the random splits (including three random splits with 0 drop) and surface-level properties of the data split. The properties' implementation is explained in detail in \cref{app:analysis_corr}.
We firstly consider \textit{task-agnostic} features: 1) the unigram overlap between the train and test set, 2) the input length in the test set and 3) the number of rare words in the test set.

Secondly, \textit{task-specific} properties are computed: 1) The number of under-represented hate keywords from the lists used by the dataset's creators (see \cref{sec:data}),
2) the number of under-represented target communities retrieved from the HateXplain annotations, and 3) a quantification of the distributional shift of data sources (Twitter and Gab are present in HateXplain) in the train and test set using the Kullback-Leibler Divergence of token distributions \citep{kullback1951information}.

\begin{table}[t]
    \centering
    \resizebox{\linewidth}{!}{\begin{tabular}{llll}
        \toprule
        & Feature & Reddit & HateXplain \\
        \midrule
        \multirow{3}{*}{task-agnostic} & unigram overlap & 0.24 & \textbf{-0.51*}\\
        & sentence length & 0.12 & 0.26 \\
        & \# Rare words & 0.13 & 0.44*\\
        \midrule
        \multirow{3}{*}{task-specific} & \# under-represented keywords & \textbf{0.47*} & 0.32*\\
        & \# under-represented targets & --- & 0.21\\
        & KL-Div. data source & --- & 0.05\\

        \bottomrule
    \end{tabular}}
    \caption{Pearson correlation between data split properties and models' F1-score drops in comparison to random splits. Correlations with a p-value $<0.05$ are marked with *. Some analysis methods are dataset-specific and cannot be computed for both datasets.}
    \label{tab:correlations_hatexplain}
    \vspace{-0.5cm}
\end{table}
\begin{figure*}[ht]
\centering
\begin{subfigure}[b]{0.47\textwidth}
   \includegraphics[width=1\linewidth]{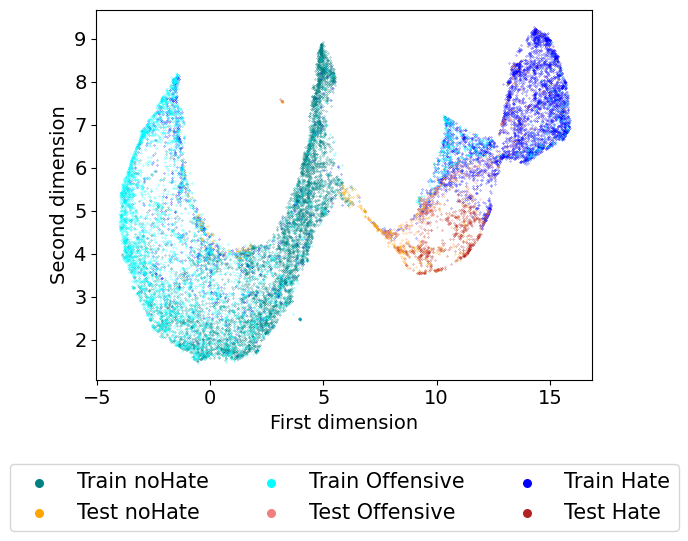}
    \caption{RoBERTa's representations} 
    \label{fig:hatexplain_hiddens_roberta}
\end{subfigure}
\begin{subfigure}[b]{0.47\textwidth}
   \includegraphics[width=1\linewidth]{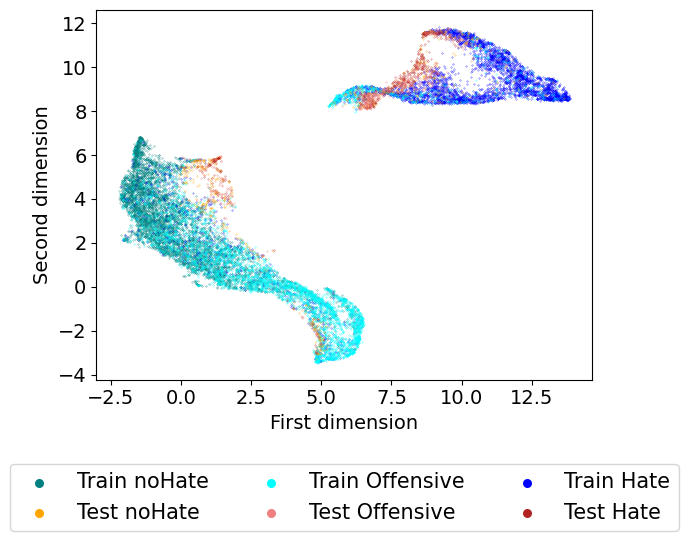}
   \caption{BERT's representations}
    \label{fig:hatexplain_hiddens_bert}
\end{subfigure}
\caption{Hidden representations for tertiary classification using the \closestsplit\ for the HateXplain dataset.}
\label{fig:hatexplain_hiddens_plot}
\vspace{-0.3cm}
\end{figure*}

\cref{tab:correlations_hatexplain} presents the results of this analysis. For the Reddit Dataset, the only significant correlation (bold) is the number of under-represented keyword categories in the training data. Task-agnostic features do not correlate with the decreased performance of models on the \closestsplit\ for the Reddit data.
In contrast, for the HateXplain dataset, task-agnostic features do play a role: The biggest (negative) correlation can be observed for the unigram overlap (bold): The higher the unigram overlap between train and test set, the closer the performance is to the random split F1-score.
Another smaller correlation exists concerning the number of rare words in the test set: The more rare words, the more challenging the split. Similar to the Reddit dataset, a significant, albeit weak, correlation exists between the decreased performance and the number of keyword categories that are under-represented in training data.

Taken together, these results suggest that the properties associated with performance drops differ from dataset to dataset. This implies that \closestsplit\ cannot easily be replicated based on task-specific or task-agnostic features. Using latent representations instead helps uncover weaknesses in models that are otherwise not easily identified.

\subsection{Visualisation of Hidden Representations}
We now take a closer look at two specific data splits for the HateXplain dataset by visualising their hidden representations. For this analysis, we select the \closestsplits\ obtained with representations with $d_B=50$ for BERT and RoBERTa, which are more commonly used than HateBERT or BERT-medium. We make these splits available via the GenBench Collaborative Benchmarking Task. 

The \closestsplit\ assigns clusters of hidden representations that are spatially close to the test set. While the clustering is conducted on high-dimensional representations, a 2-dimensional projection by UMAP \citep{mcinnes2020umap} can give an intuition about why these data splits are challenging.
\cref{fig:hatexplain_hiddens_roberta} shows RoBERTa's representations for the HateXplain dataset. A decision boundary can be observed, with mostly \textit{offensive} examples on the left, \textit{noHate} examples in the middle and \textit{hate} examples on the right. Based on this illustration, the \closestsplit\ picks a pocket of (mixed) examples between the \textit{noHate} (dark blue) and \textit{hate} (dark green) regions to be the test set.
This is mirrored in the F1-scores of the different classes. The \textit{hate} test examples lie closest to the corresponding region, and the F1-score is the highest at $47.0$. Similarly, for the \textit{noHate} class, the F1-score is relatively high at $38.28$. The offensive class, with test examples farther away, only has an F1-score of $11.88$.
The same phenomenon can be observed for a BERT-based \closestsplit\ (\cref{fig:hatexplain_hiddens_bert}).
This suggests that the model overfits its decision boundaries to train set-specific features and, therefore, fails to predict the correct classes in the test set.
Developing models using \closestsplit\ in addition to random splits might thus lead to models that are more robust to such overfitting.

\section{Conclusion}\label{sec:conclusion}
Hate speech detection systems are prone to overfitting to specific targets of hate speech and specific keywords in the input, complicating the detection of more implicit hatred and harming the generalisability to unseen demographics. Yet, in addition to those \textit{known} and \textit{interpretable} vulnerabilities, systems may have less obvious weaknesses. The data splitting method we developed aims to highlight those. Our splitting method is based on the clustering of internal representations of finetuned models, thus making the splits task- and dataset-specific. We proposed two variants (\subsetsumsplit\ and \closestsplit) that differ in how they assign clusters to the train and test set. 

The latter variant, in particular, led to consistent catastrophic drops in test set performance, when compared to a random split. 
Moreover, while each split was developed using the hidden representations from a specific model, we identified that this result generalises when developing the split using one model, and evaluating it using another.
The analyses of the resulting data splits showed that the properties of the train and test sets differ from dataset to dataset. Since no property clearly correlates with decreased model performance for both datasets, \closestsplit\ cannot be easily replicated based on data splits' surface-level properties, and using latent representations is crucial to reveal the weaknesses we observed in the models.

We encourage future work to consider evaluations using the \closestsplits\ we release for HateXplain, in order to develop more robust systems, but also emphasise that even though our results were specific to hate speech detection, the methodology can be more widely applied.
To challenge models beyond i.i.d.\ evaluation, we do not need costly data annotations. Instead, we can start by relying on systems' latent features to simulate train-test distribution shifts.
\section{Limitations}\label{sec:limitations}
We identify three main limitations of our work:
\begin{enumerate}
    \item \textbf{The scope of our work}: the splitting methodology we developed can be applied to a wide range of tasks, but we only experimented with hate speech detection.
    Future work is required to confirm the method's wider applicability.
    Moreover, even though we aim to use the challenging split to improve generalisation, we have not yet made efforts in this direction.
    \item \textbf{Generality of conclusions}: We experimented with a limited set of model architectures, all of which resemble one another in terms of their structure and the (pre-)training data used. Different models or training techniques could lead to less challenging splits, or splits with significantly different properties.
    At the same time, we did demonstrate that the split's difficulty is not model-specific (see \cref{sec:cross_model}), and observed that under variation of random seeds \closestsplit\ consistently leads to performance drops across four models and two datasets.
    \item \textbf{Naturalness of the experimental setup}: we created an artificially partitioned data split and have no guarantee that the generalisation challenges that language models encounter when deployed in real-world scenarios resemble our splits. 
    However, given that our approach simulated a worst-case scenario, demonstrated by catastrophic failure in performance, we are hopeful that models that are more robust to our train-test shift are also more robust to real-world variations in test data.
\end{enumerate}

\section{Ethics Statement}\label{sec:ethics}
By its very nature, hate speech detection involves working closely with hurtful and offensive content. This can be difficult for researchers. 
However, considering the severe consequences when hate speech models fail on unseen data and people are confronted with harmful content, it is all the more important to improve the generalisation ability of models and protect others. 

While our work intends to contribute to generalisation evaluation in a positive way, we do not recommend using our data splits as representative of generalisation behaviour `in the wild', but recommend them for academic research instead. While standard and random splits often overestimate real-world performance, our splits are likely to underestimate it, and can in this way reveal real weaknesses. 
Our splits are designed to improve academic research on the robustness of language models and contribute to improving the generalisation ability for NLP tasks.

Prior to conducting work with potentially harmful hate speech data, this project obtained approval from the Research Ethics committee at the authors' local institution.

\section*{Acknowledgments}
We thank Agostina Calabrese for helpful suggestions in the early stages of this project.
VD is supported by the UKRI Centre for Doctoral Training in Natural Language Processing, funded by the UKRI (grant EP/S022481/1) and the University of Edinburgh, School of Informatics and School of Philosophy, Psychology \& Language Sciences.
IT is supported by the Dutch National Science Foundation (NWO Vici VI.C.212.053).

\bibliography{anthology,custom}
\bibliographystyle{acl_natbib}
\clearpage
\appendix
\section{GenBench Eval Card}\label{app:eval_card}
\newcommand{\tabularwidth}{\columnwidth}

\newcommand{\expone}{$\square$}
        
\renewcommand{\arraystretch}{1.1}         
\setlength{\tabcolsep}{0mm}         
\begin{tabular}{|p{\tabularwidth}<{\centering}|}         
\hline
               
\rowcolor{gray!60}               
\textbf{Motivation} \\               
\footnotesize
\begin{tabular}{p{0.25\tabularwidth}<{\centering} p{0.25\tabularwidth}<{\centering} p{0.25\tabularwidth}<{\centering} p{0.25\tabularwidth}<{\centering}}                        
\textit{Practical} & \textit{Cognitive} & \textit{Intrinsic} & \textit{Fairness}\\
& 		
& \expone\hspace{0.8mm}		
& 		

\vspace{2mm} \\
\end{tabular}\\
               
\rowcolor{gray!60}               
\textbf{Generalisation type} \\               
\footnotesize
\begin{tabular}{m{0.17\tabularwidth}<{\centering} m{0.20\tabularwidth}<{\centering} m{0.14\tabularwidth}<{\centering} m{0.17\tabularwidth}<{\centering} m{0.18\tabularwidth}<{\centering} m{0.14\tabularwidth}<{\centering}}                   
\textit{Compo- sitional} & \textit{Structural} & \textit{Cross Task} & \textit{Cross Language} & \textit{Cross Domain} & \textit{Robust- ness}\\
& 		
& 		
& 		
& 		
& \hspace{4mm}\expone		

\vspace{2mm} \\
\end{tabular}\\
             
\rowcolor{gray!60}             
\textbf{Shift type} \\             
\footnotesize
\begin{tabular}{p{0.25\tabularwidth}<{\centering} p{0.25\tabularwidth}<{\centering} p{0.25\tabularwidth}<{\centering} p{0.25\tabularwidth}<{\centering}}                        
\textit{Covariate} & \textit{Label} & \textit{Full} & \textit{Assumed}\\  
\expone\hspace{0.8mm}		
& 		
& 		
& 		

\vspace{2mm} \\
\end{tabular}\\
             
\rowcolor{gray!60}             
\textbf{Shift source} \\             
\footnotesize
\begin{tabular}{p{0.25\tabularwidth}<{\centering} p{0.25\tabularwidth}<{\centering} p{0.25\tabularwidth}<{\centering} p{0.25\tabularwidth}<{\centering}}                          
\textit{Naturally occuring} & \textit{Partitioned natural} & \textit{Generated shift} & \textit{Fully generated}\\
& \expone\hspace{0.8mm}		
& 		
& 		

\vspace{2mm} \\
\end{tabular}\\
             
\rowcolor{gray!60}             
\textbf{Shift locus}\\             
\footnotesize
\begin{tabular}{p{0.25\tabularwidth}<{\centering} p{0.25\tabularwidth}<{\centering} p{0.25\tabularwidth}<{\centering} p{0.25\tabularwidth}<{\centering}}                         
\textit{Train--test} & \textit{Finetune train--test} & \textit{Pretrain--train} & \textit{Pretrain--test}\\
& \expone\hspace{0.8mm}		
& 		
& 		

\vspace{2mm} \\
\end{tabular}\\

\hline
\end{tabular}

Our work proposes a data split that evaluates the generalisation ability of hate speech detection models. Our motivation is an \textbf{intrinsic} one, we aim to understand better what kind of data is most challenging for hate speech detection models.

We focus on testing the \textbf{robustness} of such models, especially when it comes to out-of-distribution (o.o.d.) generalisation. However, it is not straightforward to define and detect o.o.d.\ data \citep{arora-etal-2021-types}. Moreover, data properties that might seem challenging for humans  \citep{varis-bojar-2021-sequence, ebrahimi-etal-2018-hotflip} might not be equally challenging for models or rely on costly annotations \citep{arango-2019-general, nejadgholi-kiritchenko-2020-cross, bourgeade-etal-2023-learn}. 

Therefore, we create a train test split by only relying on a model's hidden representations. This  \textbf{partitioned natural} splitting method yields a \textbf{covariate shift}, since we re-split existing data sets. The resulting train test splits indeed challenge hate speech detection models in a \textbf{finetune train-test} locus.

\newpage
\section{Clustering}\label{app:clusterh}
Our proposed data split creates a train-test split by assigning whole clusters of latent representations to either the train or the test set. We use $k$-means clustering \citep{lloyd-1982-kmeans} to perform the clustering. The used hyperparamters can be found below.
\begin{table}[!h]
\setlength{\tabcolsep}{3pt}

    \centering
    \resizebox{\linewidth}{!}{\begin{tabular}{ll}
        \toprule
        Parameter & Value \\
        \midrule
        n clusters & 3-50 \\
        n initializations with different centroids & 10 \\
        max. iterations for a run & 300 \\
        random state & 42, 62, 82 \\
        algorithm & LLoyd    \\    
        \bottomrule
    \end{tabular}}
    \caption{K-Means hyperparameters}
    \label{tab:k-means}
    \vspace{-0.3cm}
\end{table}

\section{Language Models}\label{app:lms}
We use four transformer language models to obtain and evaluate the data splits: BERT-Base(-Cased)  \citep{devlin-etal-2019-bert}, its smaller variant  BERT-Medium \citep{turc-2019-medbert, 
bhargava2021generalization}, HateBERT \citep{caselli-etal-2021-hatebert}, a BERT-Base-Uncased model that was further pretrained on abusive Reddit data using the MLM objective, and RoBERTa-Base \citep{liu-2019-roberta}. The hyperparamters for finetuning can be found below. They are generally adopted from the finetuned models from \citet{caselli-etal-2021-hatebert}, but due to computational restrictions, the models had to be trained with reduced batch sizes. To compensate for this, models were trained with more epochs with the option of early stopping.
\begin{table}[H]
\setlength{\tabcolsep}{3pt}

    \centering
    \resizebox{\linewidth}{!}{\begin{tabular}{ll}
    \toprule
    Hyperparameter & Value \\
    \midrule
         batch size & 4 (biggest possible) \\ 
         early stopping & after 5 epochs\\
         maximum epochs & 10 (20 for the larger RoBERTa models) \\
         \midrule
         optimizer & AdamW \\
         learning rate & 2e-5  \\
         adam epsilon & 1e-8 \\
         scheduling & linear schedule with warmup \\
         warm up steps & 0 \\
         \midrule
         random seeds & 42, 55, 83 \\
         max. sequence length & 512 \\
        \bottomrule
    \end{tabular}}
    \caption{Hyperparameters for finetuning the language models are adopted from the finetuned models from \citet{caselli-etal-2021-hatebert}.}
    \label{tab:lm_hyperparameters}
\end{table}

\newpage
\section{Detailed Results}\label{app:details}
The following section presents detailed results including baselines, hyperparameter selections and further results.
\subsection{Baselines}\label{app:baselines}
We compare the performance of models trained on our proposed data splits (\closestsplit\ and \subsetsumsplit) to a random split. We obtain random splits not only from 100\% of the data but also from 90\% of the data. This is necessary to compare the random split to the \closestsplit\ and \subsetsumsplit, as these use only 90\% of the data. The random split performances are presented below. 
\begin{table}[h]

\setlength{\tabcolsep}{3pt}

    \centering
        \resizebox{\linewidth}{!}{\begin{tabular}{lccc}
        \toprule
         model   &    valid acc.   & test acc.      & hate f1     \\
        \midrule
        SVM*  & -- & -- & 75.7  \\
        RNN*  & -- & -- &77.5  \\
        \midrule
         BERT-base (100\%)       & 94.6 $\pm$ 0.21  & 91.55 $\pm$ 0.13 & 82.24 $\pm$ 0.34  \\
         BERT-base (90\%)        & 91.69 $\pm$ 0.07 & 91.25 $\pm$ 0.11 & 81.96 $\pm$ 0.5   \\
         
         BERT-med.  (100\%)          & 94.3 $\pm$ 0.23  & 91.63 $\pm$ 0.2  & 82.27 $\pm$ 0.45 \\
         BERT-med.  (90\%)           & 91.84 $\pm$ 0.07 & 91.2 $\pm$ 0.15  & 81.58 $\pm$ 0.66  \\
         
         HateBert (100\%) & 94.12 $\pm$ 0.06 & 91.87 $\pm$ 0.16 & 82.72 $\pm$ 0.38\\
         HateBert (90\%)               & 92.02 $\pm$ 0.07 & 91.51 $\pm$ 0.13 & 82.34 $\pm$ 0.59 \\
         
         RoBERTa  (100\%)              & 94.4 $\pm$ 0.12  & 91.67 $\pm$ 0.2  & 82.5 $\pm$ 0.49   \\
         RoBERTa  (90\%)               &  91.8 $\pm$ 0.09  & 91.37 $\pm$ 0.16 & 82.15 $\pm$ 0.61  \\

        \bottomrule
        \end{tabular}}
    \caption{Results for the Reddit dataset on random splits using 100\% and 90\% of the data. Random splits are generated using three different seeds and models are trained with three initialisation seeds; mean and standard errors are reported. Results marked with * are taken from \citet{qian-etal-2019-benchmark}.}
    \label{tab:reddit_baselines}
\end{table}

\begin{table}[h]
    \centering
\setlength{\tabcolsep}{3pt}
\resizebox{\linewidth}{!}{\begin{tabular}{llccc}
\toprule
 split           & model   & valid acc     & test acc      & Macro f1            \\
\midrule
stand.       & BERT-base  *  &       --      &       69.0  & 67.4    \\
\midrule
  \multirow{ 4}{*}{stand.} & BERT-base        & 67.45 $\pm$ 0.36 & 68.38 $\pm$ 0.35 & 66.06 $\pm$ 0.44  \\
 &BERT-med.        & 63.93 $\pm$ 1.2  & 64.58 $\pm$ 0.99 & 62.32 $\pm$ 1.45 \\
 &HateBert               & 68.12 $\pm$ 0.16 & 68.0 $\pm$ 0.37  & 65.97 $\pm$ 0.36\\
 &RoBERTa              & 67.32 $\pm$ 0.3  & 67.83 $\pm$ 0.42 & 65.98 $\pm$ 0.26 \\
 \midrule
 \multirow{ 4}{*}{rand.}       &  BERT-base & 67.66 $\pm$ 0.31 & 68.25 $\pm$ 0.28 & 66.0 $\pm$ 0.36   \\
 &BERT-med.           & 62.46 $\pm$ 0.49 & 62.85 $\pm$ 0.42 & 60.18 $\pm$ 0.42  \\
& HateBert            & 67.91 $\pm$ 0.32 & 68.51 $\pm$ 0.28 & 66.25 $\pm$ 0.35 \\
& RoBERTa             & 66.45 $\pm$ 0.51 & 66.4 $\pm$ 0.56  & 64.1 $\pm$ 0.9    \\
\bottomrule
\end{tabular}}
    \caption{Results for the HateXplain dataset on the standard (stand.) split and on random (rand.) splits using 90\% of the data. Random splits are generated using three different seeds and models are trained with three initialisation seeds; mean and standard errors are reported. Results marked with * are taken from \citet{mathew-2020-hatexplain}. }
    \label{tab:hatexplain_baselines}
\end{table} 

\subsection{Hyperparameter Selection for Proposed Split}\label{app:split_hyper}
\begin{figure}[H]
    \centering
    \includegraphics[width=\linewidth]{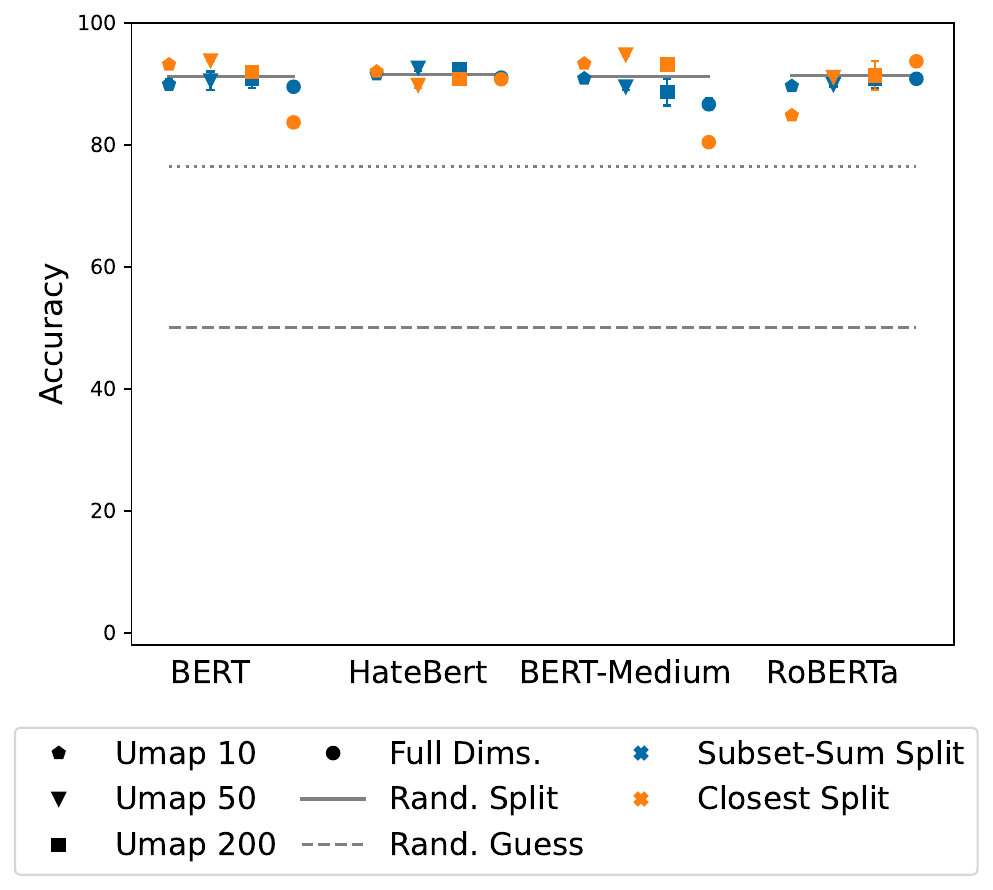}
    \caption{Performance of language models trained on the \textbf{pretrained} \subsetsumsplit\ and pretrained \closestsplit\ of the Reddit data. The errorbars show the standard error between cluster seeds.}
    \label{fig:reddit_pretrained_accuracy}
    \vspace{-0.3cm}
\end{figure}

\begin{figure}[H]
    \centering
    \includegraphics[width=\linewidth]{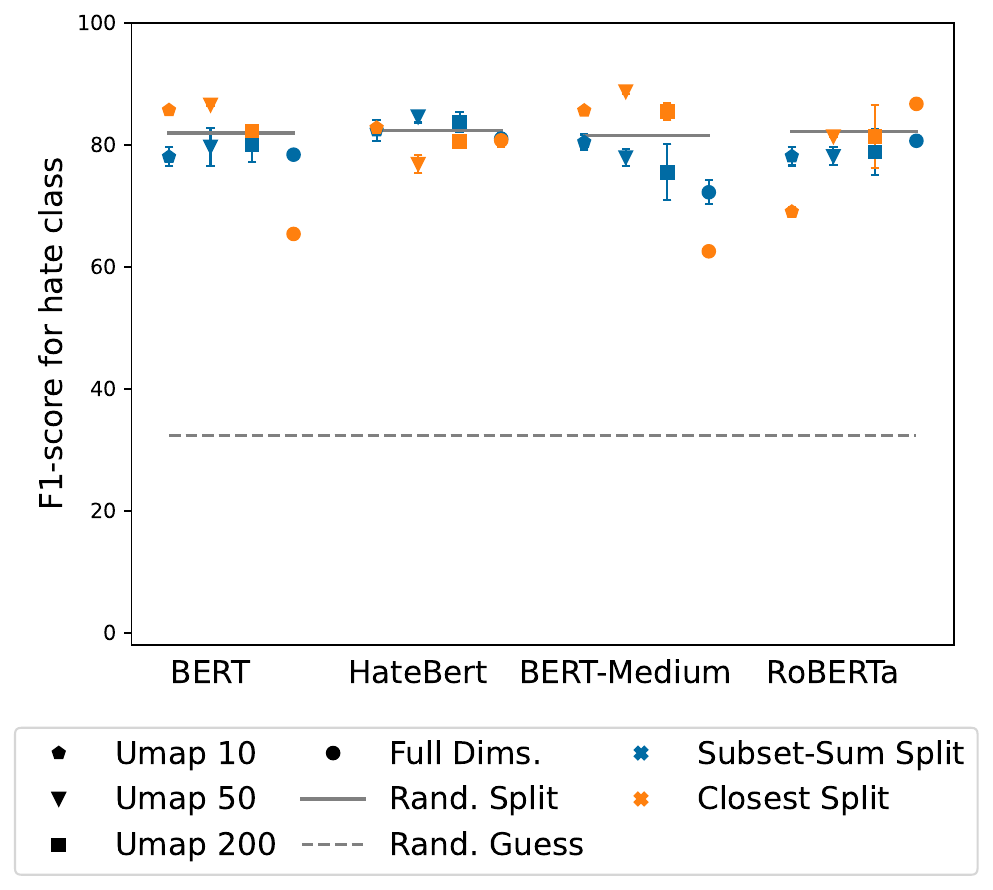}
    \caption{Performance of language models trained on the \textbf{pretrained} \subsetsumsplit\ and pretrained closest split of the Reddit data. The errorbars show the standard error between cluster seeds.}
    \label{fig:reddit_pretrained_f1}
    \vspace{-0.3cm}
\end{figure}

\begin{figure*}[ht]
\centering
\begin{subfigure}[b]{\linewidth}
    \centering
   \includegraphics[width=0.7\linewidth]{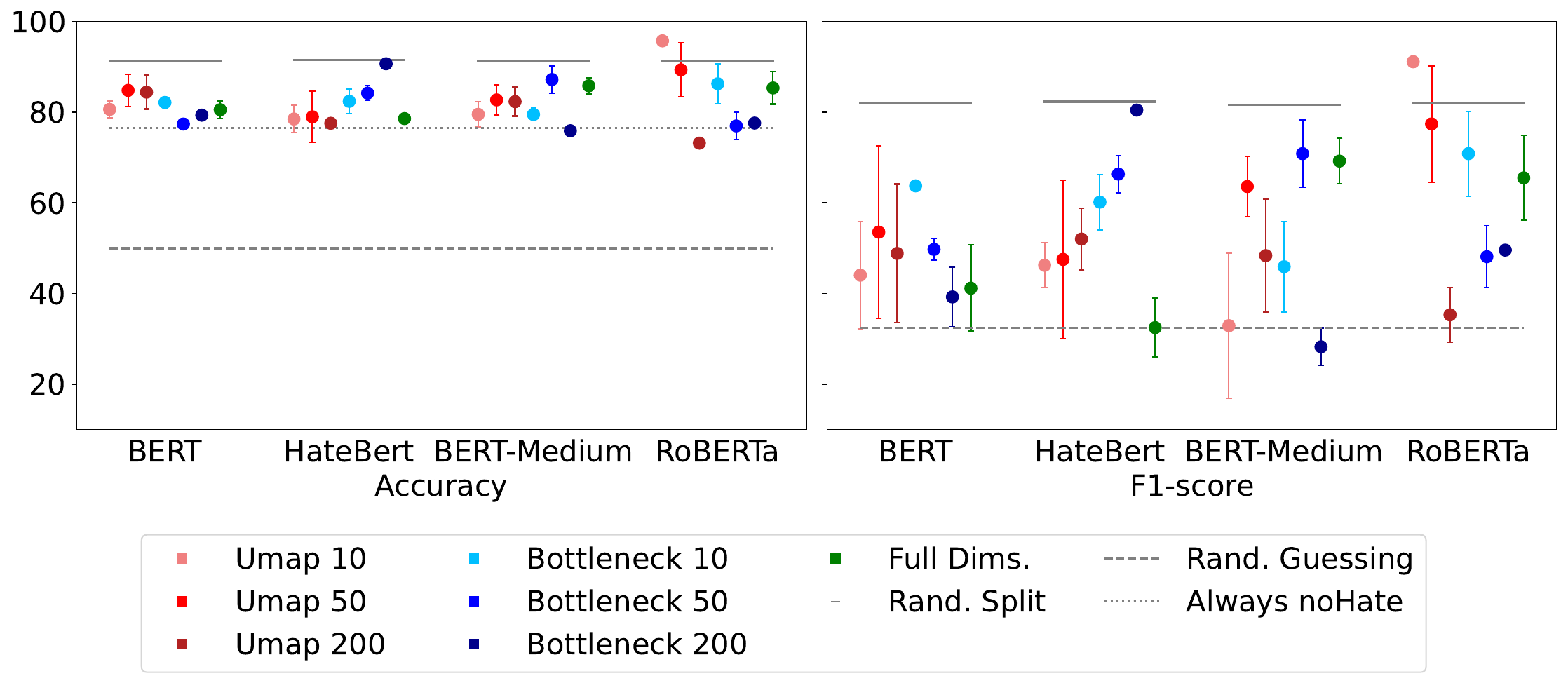}
   \caption{\subsetsumsplit}
   \label{fig:hatexplain_subset_sum} 
\end{subfigure}

\begin{subfigure}[b]{\linewidth}
    \centering
   \includegraphics[width=0.7\linewidth]{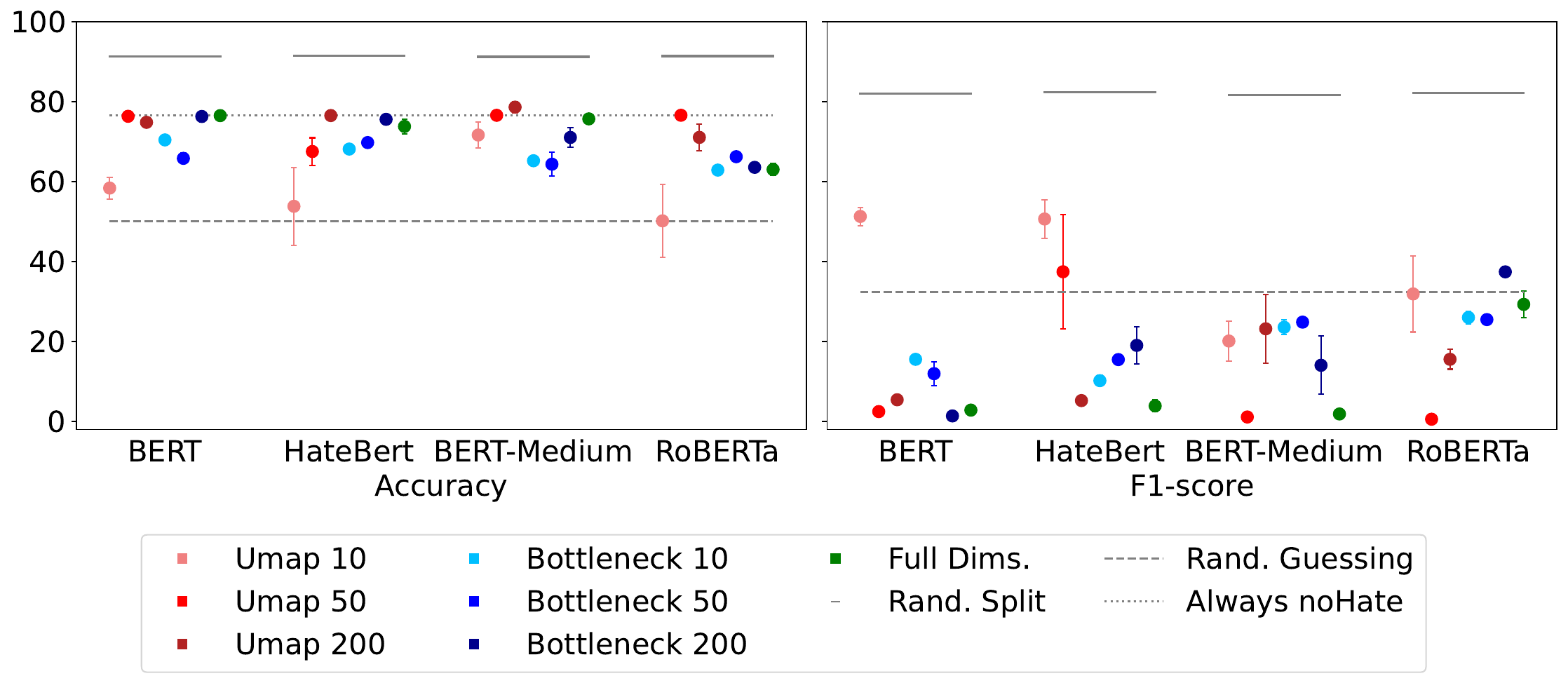}
   \caption{\closestsplit}
   \label{fig:hatexplain_closest}
\end{subfigure}

\caption{Performance of language models trained on the \subsetsumsplit\ and \closestsplit\ of the Reddit dataset. Random split performance, indicated by the solid horizontal lines, is used as a baseline. The error bars show the standard error between cluster seeds.}
\label{fig:reddit_results_more_dims}
\vspace{-0.3cm}
\end{figure*}

We analyse the effects of two hyperparameters. First, we analyse whether task-specific, finetuned representations are needed for challenging data splits or whether task-agnostic, pretrained representations also lead to difficult splits. The results can be found in \cref{fig:reddit_pretrained_accuracy} and \cref{fig:reddit_pretrained_f1}. The second hyperparameter we analyse is the dimensionality of the representations, as displayed in \cref{fig:reddit_results_more_dims}.


\subsection{Subset-Sum and Closest Split}\label{app:accs_splits}
\begin{figure}[H]
    \centering
    \includegraphics[width=\linewidth]{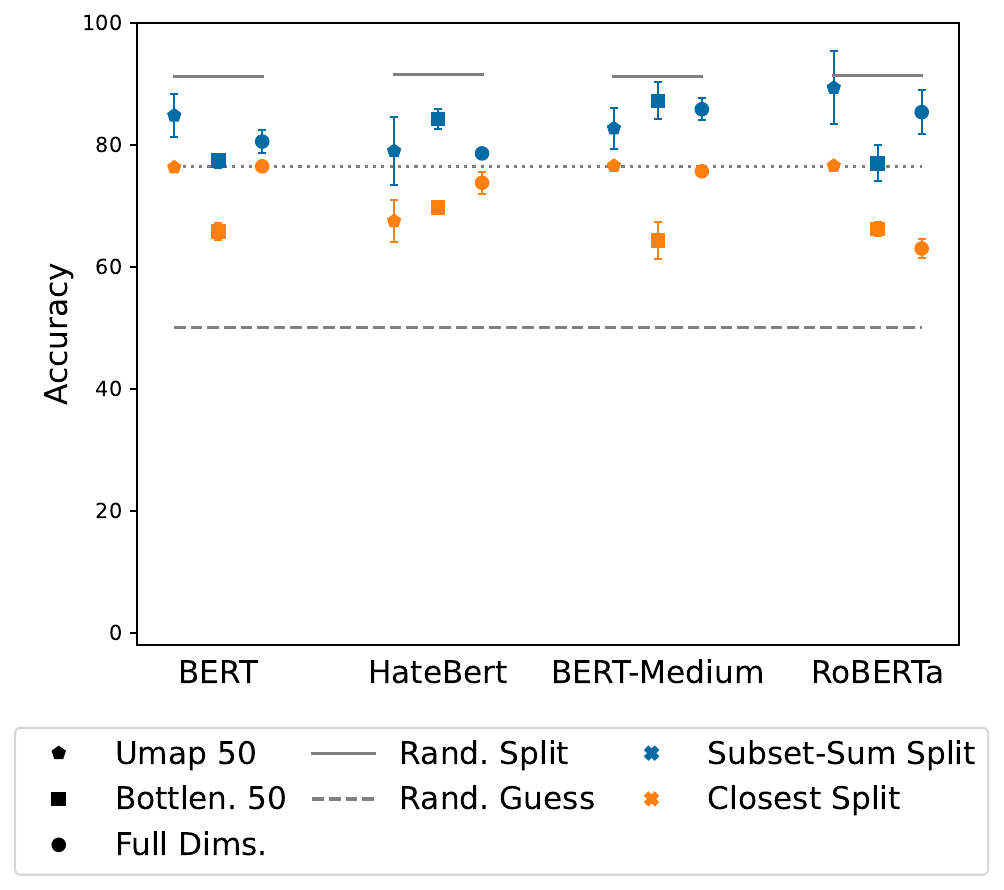}
    \caption{Performance of language models trained on the \subsetsumsplit\ and \closestsplit\ of the Reddit data. The errorbars show the standard error between cluster seeds.}
    \label{fig:reddit_accuracy}
    \vspace{-0.3cm}
\end{figure}

\subsetsumsplit\ and \closestsplit\ both lead to a decreased performance. The performance on the Reddit dataset in terms of accuracy can be found below in  \cref{fig:reddit_accuracy}.

\begin{figure}[H]
    \centering
    \includegraphics[width=\linewidth]{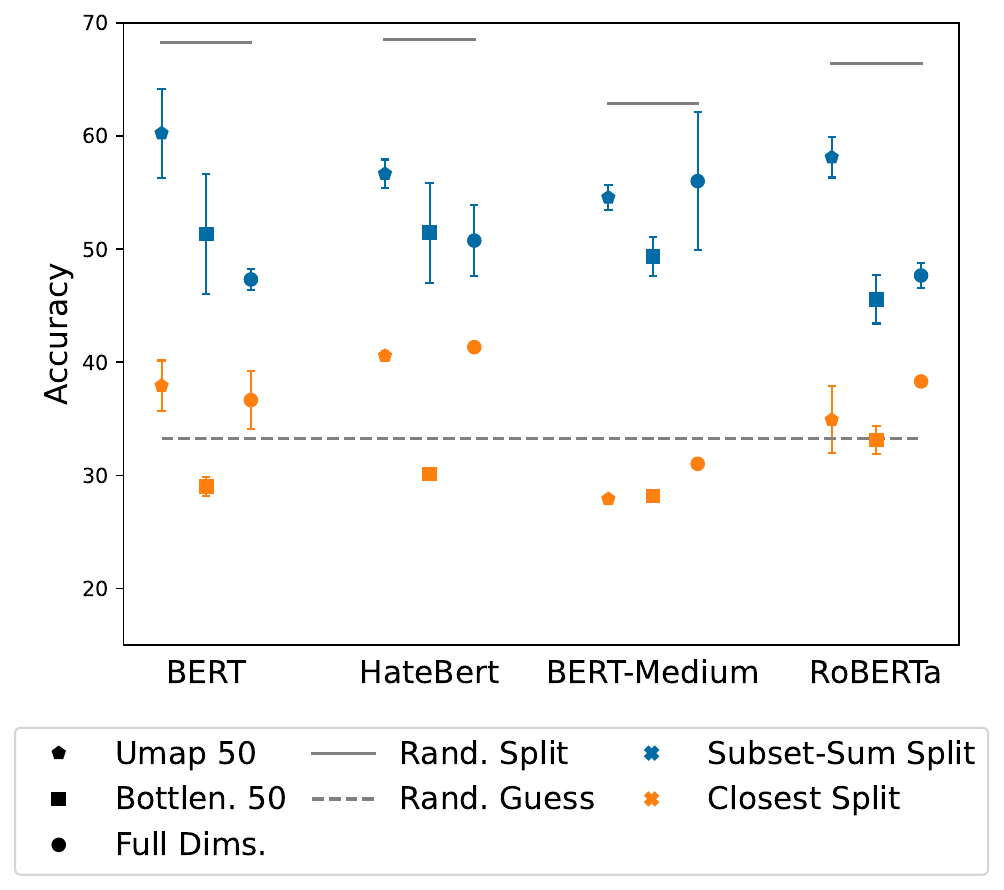}
    \caption{Performance of language models trained on the \subsetsumsplit\ and \closestsplit\ of the HateXplain data. The errorbars show the standard error between cluster seeds.}
    \label{fig:hatexplain_accuracy}
    \vspace{-0.3cm}
\end{figure}

The  HateXplain accuracy can be found in  \cref{fig:hatexplain_accuracy}. For both datasets, models  fail to predict some class completely, defaulting instead to one of the other classes. Note that HateXplain is a balanced dataset, while Reddit is highly unbalanced (75\% noHate). 

\clearpage
\newpage
\section{Analysis}\label{app:analysis}

\subsection{Data split properties}\label{app:analysis_corr}
This section presents a detailed description of the features used for the analysis in \cref{sec:analysis}. The following task-agnostic features are included in the analysis:

\paragraph{Unigram Overlap}  Following the word overlap algorithm in \citet{elangovan-etal-2021-memorization}, the word overlap $o_i$ for a given test example $test_i$ is the word overlap with the most similar training example $train_k$.  The word overlap of the whole test set is then the average over the word overlap of the test examples $o_i$. For this computation, examples are represented as a vector with unigram counts (ignoring stopwords), and similarity is computed as the cosine similarity. 

\paragraph{Sentence Length in the Test Set} We use the average length of input examples in the test set in terms of characters. 

\paragraph{Number of Rare Words in the Test Set} Rare words are defined following the definition of \citet{godbole2022benchmarking}: Rare words are words that are not common (i.e. occur at most once per million words) and are not misspelled (i.e. appear in the word list of common words\footnote{\url{https://github.com/dwyl/english-words}}). For word frequency statistics, \citet{godbole2022benchmarking} rely on \citet{Brysbaert-2009-freq}. We use the word frequencies more recently collected by \citet{robyn_speer_2022_7199437} instead.

\vspace{2mm}
\noindent Moreover, we compare the dropped performance on the proposed data splits to the following task-specific features:

\paragraph{Number of under-represented keywords in the train set}
The Reddit and HateXplain dataset have been created by filtering posts based on hate keywords by simply string-matching the posts with the keywords. These keywords can be understood as hate speech categories. 
We calculate the number of hate speech categories that are under-represented in the train set, i.e. have less than $50\%$ of their occurrences in the train set. Keywords that occur in less than 3\% of the data set are excluded.

\paragraph{Number of under-represented targets in the train set} This method aims to analyse the different targets of hate speech. For the HateXplain dataset, these targets are annotated as explained in \cref{sec:data}.  We calculate the number of under-represented targets in the train set using the same concept as for the under-represented keywords.

\paragraph{Difference of the data source distribution in the train and test set} As described in \cref{sec:data}, the HateXplain dataset consists of two data sources, Gab ($46\%$) and Twitter ($54\%)$. We calculate the distributional shift between the data source distribution in the train and test set.
The Kullback-Leibler Divergence \citep{kullback1951information} is calculated for the two data sources in the dataset and then the average is taken over both classes, weighted by the occurrence of the class in the dataset. Since there is no upper bound for the KL Divergence, it is scaled to be between $0$ and $1$ by the function
\begin{equation}
    f(x) = 1 - e^{-x}.
\end{equation}

\subsection{Topic analysis}\label{app:topics}
\setlength{\tabcolsep}{6pt}
\begin{table}[H]
    \centering
    \begin{tabular}{llll}
        \toprule
        Set & Class & Topics RoBERTa \\
     
        \midrule
        
        \multirow{3}{*}{Train} & Hate & nigger, kike, white, jews \\
        & Offens. &  retarded, bitch, white, ghetto \\ 
        & noHate & white, people, women, raped\\ 

        \midrule
        
        \multirow{3}{*}{Test}  & Hate & jews, faggot, muslim, white\\
        & Offens. & faggot, jews, nigger, white \\
        & noHate & white, jews, people, retarded \\
        \bottomrule
    \end{tabular}
    \caption{Top 4 topics for different classes in the HateXplain dataset. The topics are obtained from train and test sets of the Closest Split with latent representations from RoBERTA.}
    \label{tab:topics_hatexplain}
    \vspace{-0.3cm}
\end{table}

We extract topics for each class in the train and test sets using c-TF-IDF \citep{grootendorst2022bertopic}.

As an example, \cref{tab:topics_hatexplain} summarises the topics with the highest c-TF-IDF scores. There seems to be a tendency for the offensive and noHate classes to have different topics in the train and test sets, while the hate class is more consistent across the split. A manual analysis of cluster topics for all cluster splits did not lead to conclusive results: Topics are not clearly separated across all classes between the train and test sets. 
Many of the topics found by c-TF-IDF seem to coincide with the targets that were annotated, and used for the analysis in the previous section. No strong correlation between targets and performance was observed then, which strengthens the result that different targets in the train and test sets are not the reason for the decreased performance. 

\end{document}